\documentclass{llncs}

\usepackage{times}
\usepackage[latin1]{inputenc}

\usepackage{fadi}
\usepackage{tikz}
\usepackage{pifont}    % dingbats
\usepackage{fancybox}  % fancy boxes

\title{Opportunistic Adaptation Knowledge Discovery}
\toctitle{Opportunistic Adaptation Knowledge Discovery}
\titlerunning{Opportunistic Adaptation Knowledge Discovery}
\author{Fadi Badra\inst{1} \and Amélie Cordier\inst{2} \and Jean
  Lieber\inst{1}} 
\institute{
  LORIA (CNRS, INRIA, Nancy Universities)\\
  BP 239, 54506 Vand{\oe}uvre-l{{\`e}s}-Nancy, France\\
  \email{\{badra,lieber\}@loria.fr}\\
  \and
  LIRIS CNRS UMR 5202, Université Lyon 1, INSA Lyon,
  Université Lyon 2, ECL\\
  43, bd du 11 novembre 1918, Villeurbanne, France\\
  \email{Amelie.Cordier@liris.cnrs.fr}
}
\authorrunning{Badra et al.}
\tocauthor{
  Fadi Badra \{ LORIA \}, 
  Amélie Cordier \{ LIRIS \} and Jean
  Lieber \{ LORIA \}
}

\begin{document}

\maketitle

\begin{abstract}
Adaptation has long been considered as the Achilles' heel of
case-based reasoning since 
it requires some domain-specific knowledge that is difficult to acquire. 
In this paper, two strategies are combined in order to reduce 
the knowledge engineering cost 
induced by the adaptation knowledge ($\ca$) acquisition task: 
$\ca$ is learned from the case base by the means of knowledge
discovery techniques,
and the $\ca$ acquisition sessions are opportunistically triggered,
i.e., at problem-solving time.
\end{abstract}

%        ------------
\section{Introduction}
%        ------------
\label{sec:introduction}

% amener le sujet
Case-based reasoning ($\rapc$~\cite{DBLP:journals/aicom/MantarasP97})
is a reasoning paradigm 
based on the reuse of previous problem-solving experiences, called
cases. A $\rapc$ system often has profit of a retrieval procedure,
selecting in a case base a source case similar to the target problem,
and an adaptation procedure, that adapts the retrieved source case to
the specificity of the target problem. The adaptation procedure
depends on domain-dependent adaptation knowledge 
($\ca$, in the following).
Acquiring $\ca$ can be done from experts or by using machine learning
techniques. An intermediate approach is knowledge discovery ($\KD$)
that combines efficient learning algorithms with human-machine
interaction. 

% presenter le sujet
Most of previous $\ca$ acquisition strategies are off-line: they are
disconnected from the use of the $\rapc$ system.
By contrast, recent work aims at integrating $\ca$ acquisition from
experts to specific reasoning sessions: this \emph{opportunistic}
$\ca$ acquisition takes advantage of the problem-solving context.
%(i.e., the target problem and the retrieved source case).
%
This paper presents an approach to $\ca$ discovery that is
opportunistic: the $\KD$ is triggered at problem-solving time.

% annoncer le plan
The paper is organized as follows.
Section~\ref{sec:notionsCBR} introduces some basic notions and
notations about $\rapc$.
Section~\ref{sec:taaable} presents the $\rapc$ system $\taaable$, which
constitutes the application context of the
study, and motivates the need for adaptation knowledge acquisition in
this application context.
Section~\ref{sec:oakd} presents the proposed opportunistic and interactive $\ca$
discovery method.
In Sect.~\ref{sec:oakd:taaable}, this method is applied to 
acquire adaptation knowledge in the context of the $\taaable$ system.
Section~\ref{sec:relatedwork} discusses this approach and situates it
among related work.
Section~\ref{sec:conclusion} concludes and presents some future work.

\section{Basic Notions About $\rapc$}\label{sec:notionsCBR}

In the following, problems are assumed to be
represented in a language $\Lpb$ and solutions in a language $\Lsol$. 
A \emph{source case} represents a problem-solving episode by a pair
$(\source,\sol(\source))$, in which $\source\in\Lpb$ is 
the representation of a problem statement and $\sol(\source)\in\Lsol$ is
the representation of its associated solution. 
$\rapc$ aims at solving a \emph{target problem} $\cible$ using a set of
source cases $\basedecas$ called the \emph{case base}.
The $\rapc$ process is usually decomposed in two main steps:
retrieval and adaptation.
\emph{Retrieval} selects a source case $(\source,\sol(\source))$ from
the case base such that $\source$ is judged to be similar to $\cible$
according to a given similarity criterion.
\emph{Adaptation} consists in modifying $\sol(\source)$ in order to
propose a candidate solution $\solCand{\cible}$ for $\cible$ to the
user.
If the user validates the candidate solution $\solCand{\cible}$, then
$\solCand{\cible}$ is considered to be a solution $\sol(\cible)$ for
$\cible$. 

%<       ------------------------------------------
\section{Application Context: the $\taaable$ System}\label{sec:taaable}
%<       ------------------------------------------

%< La description du système Taaable 1 et de l'adaptation dans ce
%< système. 
The $\taaable$ system~\cite{taaable} is a cooking $\rapc$ system. 
In the cooking domain, $\rapc$ aims at answering a query using a set
of recipes.  
In order to answer a query, the system retrieves a recipe in the recipe
set and adapts it to produce a recipe satisfying the query.
The $\taaable$ system was proposed to participate to the
\emph{Computer Cooking Contest} (CCC) challenge in
2008~\cite{DBLP:conf/ewcbr/2008w}. 
In the CCC challenge, queries are given in natural language 
and express a set of constraints that 
the desired recipe should satisfy.
These constraints concern 
the ingredients to be included or avoided, 
the type of ingredients (e.g., meat or fruit),
the dietary practice (e.g., nut-free diet), 
the type of meal (e.g., soup)
or the type of cuisine (e.g., chinese cuisine).
An example of query is: 
``Cook a chinese soup with leek but no
  peanut oil.''
Recipes are given in textual
form, with a shallow XML structure,
and include a set of ingredients
together with a textual part describing the recipe preparation.
The $\taaable$ system is accessible online (\url{http://taaable.fr}). 

%           ---------------------
\subsection{Representation Issues}
%           ---------------------
\paragraph{A Cooking Ontology.}
The system makes use of a cooking ontology $\Ontologie$ 
represented in propositional logic.
Each concept of $\Ontologie$ corresponds to a propositional
variable taken from a finite set $\vocabulaire$ of
propositional variables. 
$\Ontologie$ is mainly composed of a 
set of concepts organized in a hierarchy, which corresponds, in
propositional logic, to a set of logical implications $\fm{a}
\limplique \fm{b}$. 
For example, the axiom $\fm{leek} \limplique \fm{onions}$ of
$\Ontologie$ states that leeks are onions.

\paragraph{Problem and Solution Representation.}
In $\taaable$, a problem $\prb\in\Lpb$ represents a query and a
solution $\sol(\prb)$ of $\prb$ represents a recipe that matches this
query. 
$\Lpb$ and $\Lsol$ are chosen fragments of propositional logic defined
using the vocabulary $\vocabulaire$ introduced in the cooking ontology
$\Ontologie$. 
One propositional variable is defined in $\Lpb$ and $\Lsol$ 
for each concept name of $\Ontologie$ and the only
logical connective used in $\Lpb$ and $\Lsol$ is the conjunction
$\land$.
For example, the representation $\cible\in\Lpb$ of
the query mentioned above is:  
$$
\cible = \fm{chinese} \land \fm{soup} \land \fm{leek} \land
\neg\,\fm{peanut\_oil}  
$$

The case base $\basedecas$ contains a set of recipes.
Each recipe is indexed in the case base by a propositional
formula $\recette\in\Lsol$.
For example, the index $\recette$ of the recipe \emph{Wonton Soup} is:
$$\recette = \fm{chinese}
\land \fm{soup} \land \fm{green\_onion} \land \ldots \land
\fm{peanut\_oil}  \land \text{\emph{Nothing else}}$$ 
\emph{Nothing else} denotes a conjunction of negative literals
$\not\fm{a}$ for all $\fm{a}\in\vocabulaire$ such that  
$ \fm{chinese}
\land \fm{soup} \land \fm{green\_onion} \land \ldots \land
\fm{peanut\_oil} \nDecouleOntologie \fm{a}$.
This kind of ``closed world assumption'' states explicitly that for all
propositional variable $\fm{a}\in\vocabulaire$, either $\recette
\DecouleOntologie \fm{a}$ (the recipe contains the ingredient
represented by $\fm{a}$) or $\recette \DecouleOntologie \not\fm{a}$
(the recipe does not contain the ingredient represented by $\fm{a}$).

Each recipe index $\recette$ represents a set of
source cases: 
$\recette$ represents the set of
source cases $(\source,\sol(\source))$ such that 
$\sol(\source) = \recette$ and $\source$ is solved by
$\recette$, 
i.e., $\source$ is such that $\recette \DecouleOntologie \source$.

%           --------------------
\paragraph{Adaptation Knowledge.}
%           --------------------
% ak = reformulations
In $\taaable$, adaptation knowledge is given by a set of
reformulations $(\pbreform,\solreform)$ in which $\pbreform$ is a binary
relation between problems and $\solreform$ is an adaptation function
associated with $\pbreform$~\cite{reformulations}.
A reformulation has the following semantics: 
if two problems $\prb_1$ and $\prb_2$ are related by 
$\pbreform$ ---denoted by $\prb_1 \mathrel{\pbreform} \prb_2$---
then for every recipe $\sol(\prb_1)$ matching the query $\prb_1$, 
$\operatorname{\solreform}(\prb_1,\sol(\prb_1),\prb_2) =
\solCand{\prb_2}$ matches the query $\prb_2$.  

In this paper, binary relations $\pbreform$ are given by substitutions
of the form $\sigmaQ = \alpha \substituepar \beta$, where $\alpha$ and
$\beta$ are literals (either positive or negative).
For example, the substitution $\sigma = \fm{leek} \substituepar
\fm{onions}$ generalizes leek into onions.

Adaptation functions $\solreform$ are given by substitutions of the form
$\sigmaR = \fm{A} \substituepar  
\fm{B}$ in which $\fm{A}$ and $\fm{B}$ are conjunctions of literals.
For example, the substitution 
$\sigmaR = \fm{soup} \land \fm{pepper} \substituepar \fm{soup} \land
\fm{ginger}$ 
states that pepper can be replaced by ginger in soup recipes.
A substitution $\sigmaR$ can be automatically generated
from a substitution $\sigmaQ$:
$\sigmaR = \fm{b} \substituepar \fm{a}$ if $\sigmaQ$
is of the form $\fm{a} \substituepar \fm{b}$ and
$\sigmaR = \emptyset \substituepar \not\fm{a}$ if $\sigmaQ$ is of the
form $\not\fm{a} \substituepar \emptyset$.

The main source of adaptation knowledge is the ontology $\Ontologie$.
A substitution $\sigmaQ = \fm{a} \substituepar \fm{b}$ is
automatically generated from each axiom $\fm{a} \limplique \fm{b}$ of
$\Ontologie$ and correspond to a \emph{substitution by generalization}.
A substitution $\sigmaQ = \fm{a} \substituepar \fm{b}$ can be applied to a
query $\prb$ if $\prb \DecouleOntologie \fm{a}$.
$\sigmaQ$ generates a new query $\sigmaQ(\prb)$ in which the
propositional variable $\fm{a}$ has been 
substituted by the propositional variable $\fm{b}$. 
For example, the substitution $\sigmaQ = \fm{leek} \substituepar
\fm{onions}$ is 
generated automatically from the axiom $\fm{leek} \limplique
\fm{onions}$ of $\Ontologie$.
$\sigmaQ$ can be applied to the query $\cible$ to produce the query 
$
\sigmaQ(\cible) = \fm{chinese} \land \fm{soup} \land \fm{onions}
\land \neg\,\fm{peanut\_oil}
$, in which $\fm{leek}$ has been 
substituted by $\fm{onions}$.
For each propositional variable $\fm{a}$ of $\vocabulaire$, 
an additional substitution of the form $\sigmaQ = \neg\,\fm{a} \substituepar
\emptyset$ is generated. 
Such a substitution can be applied to a problem $\prb$ if $\prb
\DecouleOntologie \not\fm{a}$ and generates a new problem
$\sigmaQ(\prb)$ in which the negative literal $\not\fm{a}$ is
removed. 
This has the effect to loosen the constraints imposed on a query e.g., by
omitting in the query an unwanted ingredient.
For example, the substitution $\not\fm{peanut\_oil} \substituepar
\emptyset$ applied to $\cible$ generates the query
$\sigmaQ(\cible) = \fm{chinese} \land \fm{soup} \land \fm{leek}$,
in which the condition on the ingredient $\fm{peanut\_oil}$ is omitted.

However, when $\Ontologie$ is the only source of adaptation
knowledge, the system is only able to perform simple adaptations, in
which the modifications made to 
$\sol(\source)$ correspond to a sequence of substitutions that can be
used to transform $\source$ into $\cible$.
Therefore, an additional adaptation knowledge base $\akb$ is
introduced.
$\akb$ contains a set of reformulations $(\sigmaQ,\sigmaR)$ that
capture more complex adaptation strategies.

%           ---------------------------------
\subsection{The $\rapc$ Process in $\taaable$}\label{subsec:taaableCBR}
%           ---------------------------------

% retrieval 
\paragraph{Retrieval.}
The retrieval algorithm is based on a \emph{smooth
  classification} algorithm on an index hierarchy. %~\cite{lieber02}.
Such an algorithm aims at determining a set of modifications to apply
to $\cible$ in order to obtain a modified query $\source$ that
matches at least one recipe $\sol(\source)$ of the case base.
The algorithm computes a \emph{similarity path}, which is a
composition of substitutions  
$\simPath =
\sigmaQ_q\comp\sigmaQ_{q-1}\comp\dotsb\comp\sigmaQ_{1}$   
such that there exists at least one recipe $\sol(\source)$
matching the modified query  
$\source = \sigmaQ_q(\sigmaQ_{q-1}(\ldots\sigmaQ_1(\cible)\ldots))$,
i.e., such that $\sol(\source) \DecouleOntologie \source$ holds. 
Thus, a similarity path $\simPath$ can be written:
$$
\sol(\source) \DecouleOntologie \source \xleftarrow{\sigmaQ_q}  
\xleftarrow{\sigmaQ_{q-1}} \dotsb \xleftarrow{\sigmaQ_{1}} \cible 
$$
For example, to solve the above query $\cible$, the system
generates a similarity path $\simPath =
\sigmaQ_2\comp\sigmaQ_1$, with:
\begin{align*}
\cible &= \fm{chinese} \land \fm{soup} \land \fm{leek} \land
\not\fm{peanut\_oil} \\ 
\sigmaQ_1 &= \neg\,\fm{peanut\_oil} \substituepar \emptyset,
\quad\sigmaQ_2 = \fm{leek} \substituepar \fm{onions}\\
\source &= \fm{chinese} \land \fm{soup} \land \fm{onions}\\
\sol(\source) &= \fm{chinese}
\land \fm{soup} \land \fm{green\_onion} \land \ldots \land
\fm{peanut\_oil}  \land \text{\emph{Nothing else}}
\end{align*}
In this similarity path, 
$\sol(\source)$ is the propositional representation of the recipe
\emph{Wonton Soup}. 
Since the ontology $\Ontologie$ contains the axiom
$\fm{green\_onion} \limplique \fm{onions}$,
the modified query $\source=\sigmaQ_2\mathrel{\circ}\sigmaQ_1\,(\cible)$ 
verifies $\sol(\source) \DecouleOntologie \source$. 

% adaptation
\paragraph{Adaptation.}
To a similarity path is associated an \emph{adaptation path}
$\adaptationPath$, which is a composition
of substitutions 
$\adaptationPath =
\sigmaR_1\comp\sigmaR_{2}\comp\dotsb\comp\sigmaR_{q}$ such that 
the modified recipe $\solCand{\cible} =
\sigmaR_1(\sigmaR_{2}(\ldots\sigmaR_q(\sol(\source))\ldots))$ solves the
initial query $\cible$, i.e.,  
verifies $\solCand{\cible} \DecouleOntologie \cible$.
Thus, an adaptation path $\adaptationPath$ can
be written
$$
\sol(\source) \xrightarrow{\sigmaR_q} \xrightarrow{\sigmaR_{q-1}} \dotsb
\xrightarrow{\sigmaR_1} \solCand{\cible} \DecouleOntologie \cible
$$

The adaptation path $\adaptationPath$ is constructed from the
similarity path $\simPath$ by associating a substitution $\sigmaR_i$
to each substitution $\sigmaQ_i$. 
To determine which substitution $\sigmaR_i$
to associate to a given substitution $\sigmaQ_i$,
the external adaptation knowledge base $\akb$ is searched first.
For a substitution $\sigmaQ_i = \alpha \substituepar \beta$, 
the system looks for a substitution $\sigmaR = \fm{A} \substituepar
\fm{B}$ such that $\fm{A} \DecouleOntologie \beta$ and $\fm{B}
\DecouleOntologie \alpha$.
For example, if 
$\sigmaQ_2 = \fm{leek} \substituepar \fm{onions}$ is used in $\simPath$
and $\akb$ contains the reformulation $(\sigmaQ,\sigmaR)$ with
$\sigmaQ = \sigmaQ_2$ and $\sigmaR =
\fm{green\_onion} \substituepar \fm{leek} \land \fm{ginger}$,
$\sigmaR$ will be selected to constitute the substitution $\sigmaR_2$
in $\adaptationPath$ since $\fm{green\_onion} \DecouleOntologie
\fm{onions}$ and $\fm{leek} \land \fm{ginger} \DecouleOntologie \fm{leek}$.
If no substitution $\sigmaR$ is found in $\akb$ for a given
substitution $\sigmaQ_i$ then $\sigmaR_i$ is generated automatically
from $\sigmaQ_i$.

\begin{figure}[h]
\begin{center}
\begin{tikzpicture}
  \path
  (0,2) node[rectangle] (source) { $\source$ }
  (0,0) node[rectangle] (solsource) { $\sol(\source)$ }
  (3,2) node[rectangle] (pb) { $\prb$ }
  (3,0) node[rectangle] (solpb) { $\solCand{\prb}$ }
  (6,2) node[rectangle] (cible) { $\cible$  }
  (6,0) node[rectangle] (solcible) { $\solCand{\cible}$ };
  \draw[-,black,thick] (cible.west) -- node[midway,above]
       {$\sigmaQ_1$} (pb.east); 
  \draw[-,black,thick] (pb.west) -- node[midway,above] {$\sigmaQ_2$}
  (source.east); 
  \draw[-,black,thick] (solsource.east) -- node[midway,below]
       {$\sigmaR_2$} (solpb.west);  
  \draw[-,black,thick] (solpb.east) -- node[midway,below]
       {$\sigmaR_1$} (solcible.west);  
  \draw[->,black,thick] (source) -- (solsource);
  \draw[->,black,thick] (pb) -- (solpb);
  \draw[->,black,thick] (cible) -- (solcible);
\end{tikzpicture}
\end{center}
\caption{A similarity path and the associated adaptation path.}
\label{fig:reformulations}
\end{figure}
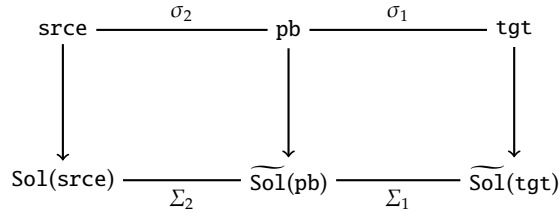

In the previous example, $\akb$ is considered to be empty so $\sigmaR_1$ and
$\sigmaR_2$ are generated automatically from the substitutions
$\sigmaQ_1$ and $\sigmaQ_2$:
 $\sigmaR_1 = \emptyset \substituepar \not\fm{peanut\_oil}$
since $\sigmaQ_1 = \not\fm{peanut\_oil} \substituepar \emptyset$ and 
$\sigmaR_2 = \fm{onions} \substituepar \fm{leek}$ 
since $\sigmaQ_2 = \fm{leek} \substituepar \fm{onions}$.
According to the axiom $\fm{green\_onion} \limplique \fm{onions}$ of
$\Ontologie$, the system further specializes the substitution
$\sigmaR_2$ into the substitution $ \fm{green\_onion} \substituepar 
\fm{leek}$ and the user is proposed to replace green onions by
leek in the recipe \emph{Wonton Soup} and to suppress peanut oil.
The generated adaptation path is $\adaptationPath =
\sigmaR_1\comp\sigmaR_2$~(Fig.~\ref{fig:reformulations}), with:  
\begin{align*}
\sol(\source) &= \fm{chinese}
\land \fm{soup} \land \fm{green\_onion} \land \ldots \land
\fm{peanut\_oil}  \land \text{\emph{Nothing else}} \\
\sigmaR_2 &= \fm{green\_onion} \substituepar \fm{leek},
%% \solCand{\prb} &= \fm{chinese}
%% \land \fm{soup} \land \fm{leek} \land \ldots \land
%% \fm{peanut\_oil}  \land \text{\emph{Nothing else}}\\
\quad\sigmaR_1 = \emptyset \substituepar \not\fm{peanut\_oil} \\
\solCand{\cible} &= \fm{chinese}
\land \fm{soup} \land \fm{leek} \land \ldots \land
\not\fm{peanut\_oil}  \land \text{\emph{Nothing else}}\\
\cible &= \fm{chinese} \land \fm{soup} \land \fm{leek} \land
\not\fm{peanut\_oil} 
\end{align*}
The inferred solution $\solCand{\cible}$ solves the initial query
$\cible$: $\solCand{\cible} \DecouleOntologie \cible$.

%           ------------------------------------------------
\subsection{Why Learning Adaptation Knowledge in $\taaable$?}
%           ------------------------------------------------
In the version of the $\taaable$ system that was proposed to
participate in the CCC challenge, $\akb = \emptyset$ so
adaptation knowledge is inferred from the ontology
$\Ontologie$. 
The main advantage of this approach lies in its simplicity: no
external source of adaptation knowledge is needed and the system is
able to propose a solution to any target problem.
However, the system's adaptation capabilities (simple substitutions)
appear to be very limited and the user has no means to give some
feedback on the quality of the proposed adaptation.

For example, the substitution $\sigmaR_1 = \emptyset \substituepar
\not\fm{peanut\_oil}$ suggests to remove the
ingredient peanut oil in the retrieved recipe, but as the oil is used
in this recipe to saute the bok choy, the adapted recipe turns out to
be practically unfeasible.
A better adaptation would suggest to replace peanut oil by e.g.,
sesame oil, which can be modeled by the substitution 
$\sigmaR_1 = \fm{peanut\_oil} \substituepar \fm{sesame\_oil}$.
To generate this substitution automatically, the system could for
example exploit the fact that the concepts 
$\fm{peanut\_oil}$ and $\fm{sesame\_oil}$ are both sub-concepts of the
concept $\fm{oil}$ in $\Ontologie$. 
But still, some additional knowledge would be needed to
express the fact that peanut oil should be replaced by sesame oil, and
not by olive oil or hot chili oil, as $\fm{olive\_oil}$ and
$\fm{hot\_chili\_oil}$ are also  
sub-concepts of $\fm{oil}$ in $\Ontologie$.
Besides, the system should be aware that this substitution is recommended
\emph{only in Asian cuisine}, which can be modeled by the more precise
substitution  
$\sigmaR_1 = \fm{asian} \land \fm{peanut\_oil} \substituepar
\fm{asian} \land \fm{sesame\_oil}$.

Furthermore, the second substitution 
$\sigmaR_2 = \fm{green\_onions} \substituepar \fm{leek}$
suggests to solely replace sliced green onions by uncooked leek.
But the green onion was used in the original \emph{Wonton Soup}
for garniture, so the user might consider that raw leek added as garniture
alters too much the taste of a soup.
A better adaptation would consist in frying leek with
e.g., tempeh and red bell pepper to prepare the garniture.
Such an adaptation can be modeled by the substitution
$\sigmaR_2 = \fm{green\_onions} \substituepar \fm{leek} \land \fm{tempeh} \land
\fm{red\_bell\_pepper}$.
This substitution, which reflects a cooking know-how, can hardly be
generated automatically from the ontology.

These examples show that in order to improve its adaptation
capabilities, the system would greatly 
benefit from the availability of a set of adaptation rules that would
capture more complex adaptation strategies.
These adaptation rules cannot be generated automatically from the
ontology and need to be acquired from other knowledge sources.
These examples also show that the human expert plays a major role in
adaptation knowledge acquisition and that in the cooking domain,
adaptation rules are often highly contextual.
   
%<       --------------------------------------------
\section{Opportunistic Adaptation Knowledge Discovery}\label{sec:oakd}
%<        --------------------------------------------

The presented $\ca$ acquisition method combines two previous
approaches of $\ca$ acquisition.
%%  that were implemented in the
%% $\cabamaka$ system~\cite{ijcai07} and in the $\iaka$
%% system~\cite{cordier2008d}. 
%% The $\cabamaka$ system learns $\ca$ from differences between cases by
%% the means of 
%% knowledge discovery techniques (section~\ref{subsec:akd}), while
%% the $\iaka$ system acquires adaptation knowledge at problem-solving time through
%% interactions with the user (section~\ref{subsec:oka}).
The first one was implemented in the $\cabamaka$ system~\cite{ijcai07}
and learns $\ca$ from differences between cases by the means of
knowledge discovery techniques (section~\ref{subsec:akd}).
The second one was implemented in the $\iaka$ system~\cite{cordier2008d} and 
acquires adaptation knowledge at problem-solving time through
interactions with the user (section~\ref{subsec:oka}).

%<          ----------------------------------
\subsection{Adaptation Knowledge Discovery from the Case
  Base}\label{subsec:akd} 
%<          ----------------------------------

% knowledge discovery
Machine learning algorithms aim at extracting some regularities from a
set of observations.
Knowledge discovery techniques combine efficient machine 
learning algorithms with human-machine interaction.
In \cite{ijcai07}, $\ca$ is learned from differences between cases by
the means of knowledge discovery techniques.  
A set of pairs of sources cases is taken as input of a
frequent itemset extraction algorithm, which outputs a set of
itemsets. Each of these itemsets can be interpreted as an adaptation rule.
This approach of $\ca$ learning was motivated by the original idea
proposed by Kathleen Hanney and Mark T. Keane in \cite{hanney97}, in
which the authors suggest that $\ca$ may be learned from
differences between cases.
The main assumption is that the
differences that occur between cases in the case base 
are often representative of differences that will occur between future
problems and the case base.

% representing variations
To learn adaptation rules from differences between cases,
representing variations between cases is essential.
In~\cite{stairs}, expressive representation formalisms are proposed
and it is shown 
that defining a partial order on the variation language can help
organizing the learned rules by generality.

%           ---------------------------------------------------
\subsection{Opportunistic and Interactive Knowledge Acquisition}\label{subsec:oka}
%           ---------------------------------------------------

Experiential knowledge, or know-how, can often be acquired on-line, when
users are using $\rapc$ tools.
It is the aim of interactive and opportunistic knowledge acquisition
strategies to support such an acquisition.
In these strategies, the system exploits its interactions with its user to
build new pieces of knowledge, to test them and, in case of success, to
retain them.
Moreover, the knowledge acquisition process is often opportunistic, i.e,
triggered by a previous reasoning failure:
reasoning failures highlight missing knowledge and thus constitute a
guidance for the acquisition process.
A major advantage of interactive knowledge acquisition strategies is that
they ensure that the user is in a
favorable context when he participates to the acquisition process.
In~\cite{cordier2008e}, a review of interactive and opportunistic knowledge
acquisition approaches is proposed, and two strategies are developed.
This work illustrates the efficiency of interactive and opportunistic
knowledge acquisition approaches to acquire specific knowledge.
On the other hand, it shows that such approaches only allow the systems to
acquire small pieces of knowledge at a time.

%% In these approaches, knowledge acquisition is performed during
%% the reasoning process, when the expert is using the system for the
%% problem solving purpose: acquired knowledge units are "co-built" by
%% the expert and the system through interactions.
%% Interactive knowledge acquisition is often triggered in an
%% opportunistic manner, i.e., when a reasoning failure occurs.
%% Reasoning failures highlight missing knowledge in the system and help
%% guiding the acquisition process.
%% Hence, interactive and opportunistic knowledge acquisition offers an
%% alternative way to grasp knowledge "in context". 

%           ----------------------------
\subsection{Combining the two Approaches}
%           ----------------------------

%% Avantages de l'akd
When properly used, knowledge discovery techniques may have
the strong advantage of automating a part of the knowledge acquisition
process. 
% #1 - assistance
In these approaches, dedicated human-machine interfaces allow the
expert, through predefined interactions, to provide feedback on a set
of suggestions generated automatically by the system. 
The role of the expert is thus reduced to the validation of a pre-selected
set of knowledge pieces.
The acquired knowledge is directly usable by the system, without the
need for an additional formalization step. 
% #2 - ML
Automatic approaches also benefit from efficient machine learning
algorithms that can be applied, as in~\cite{stairs}, to learn 
adaptation rules at different levels of generality.
However, these approaches still produce a large number of candidate
knowledge units that have to  be validated by a domain expert out of
any context, which constitutes an important drawback.

%% Inconvénients de l'akd offline
Acquiring adaptation knowledge \emph{offline}, i.e., independently
of a particular problem-solving session, appears to be problematic.
% #1 - need for anticipation
Offline $\ca$ acquisition forces the system's designer to anticipate
the need for adaptation knowledge in problem-solving and to acquire it
in advance, which can be very tedious, if not impossible. 
% #2 - contextual rules
Offline acquisition of adaptation knowledge also makes difficult to
come up with fine-grained adaptation rules, since adaptation knowledge
is often highly contextual. 
For example, in the cooking domain, an egg
can sometimes be substituted by 100 grams of tofu, but
this adaptation rule may be applied only to certain types
of dishes, like cakes or mayonnaise, and has proved to
be irrelevant in order to adapt a mousse recipe or an
omelet recipe.
Acquiring such a rule would require to circumscribe its domain of
validity in order to avoid over-generalization.

Moreover, initial acquisition of adaptation knowledge prevents the
system from learning from experience.
A $\rapc$ system with fixed adaptation knowledge has no way to
improve its problem-solving capabilities, except by retaining in the
case base a new experience each time a problem has been
solved, as it is usually done in traditional $\rapc$
systems~\cite{DBLP:journals/aicom/MantarasP97}. 

On the other hand, interactive and opportunistic knowledge acquisition
approaches heavily rely on the human expert but ensure that the expert
is ``in context'' when validating 
knowledge units that are to be acquired.
Combining knowledge discovery techniques and interactive approaches,
as it is proposed here, could overcome one of the limitations of KD by
dramatically reducing the number of candidate adaptation rules
presented to the expert. 
By triggering the process in an opportunistic manner, the expert is able
 to parametrize the KD in order to focus on specific knowledge to acquire
 in context.
The resulting $\ca$ discovery process: 
\begin{itemize}
\item is performed \emph{on-line}, i.e., in the context of a problem-solving
  session,
\item is \emph{interactive} as adaptation knowledge is learned by the
  system through interactions with its user who acts as an expert,
\item is \emph{opportunistic} as it is triggered by reasoning failure, and,
consequently, often helps repairing a failed adaptation,
\item makes use of knowledge discovery techniques to provide
  \emph{assistance} to the user in the formulation of new knowledge:
  the user is presented with a set of suggestions that are generated
  automatically from the case base.
\end{itemize}

%<       --------------------------------------------
\section{Applying Opportunistic $\ca$ Discovery to
  $\taaable$}\label{sec:oakd:taaable} 
%< l'oakd appliqué à Taaable !
%<        --------------------------------------------

%< 1. Rememoration 
In this section, an opportunistic $\ca$ discovery process is
applied to the context of the $\taaable$ system.
  
%           ---------------
\subsection{$\ca$ Discovery}
%           ---------------

In $\taaable$, the $\ca$ discovery process consists in learning
a set of substitutions from the case base by comparing two
sets of recipes. 
 
%          -----------------
\paragraph{The Training Set.}
%          -----------------
The training set $\trainingset$ is formed by selecting from the case
base a set of pairs of recipes
$(\recette_k,\recette_\ell)\in\basedecas\times\basedecas$ and by 
representing for each selected pair of recipes
$(\recette_k,\recette_\ell)$ the   
variation $\dRkl$ from $\recette_k$ to $\recette_\ell$.
The choice of the training set 
$\trainingset$ results from a set of interactions with the user 
%is performed in interaction with the user.
during which he/she is asked to formulate the cause of the
adaptation failure and to pick up a repair strategy.
%% Thus, $\trainingset$ is determined in an opportunistic manner and in
%% interaction with the user.

%           -----------------------
\paragraph{Representing Variations.}
%           -----------------------
The variation $\dRkl$ from a recipe $\recette_k$ to a recipe 
$\recette_\ell$ is represented in a language $\LvarR$ by a
set of properties.
Three properties $\fm{a}\Moins$, $\fm{a}\Plus$ and $\fm{a}\Eg$ are
defined in $\LvarR$ for each propositional variable $\fm{a}$ of
$\vocabulaire$, and $\dRkl\in\LvarR$ contains: 
\begin{itemize}  
\item  the property $\fm{a}\Moins$ 
if $\recette_k\DecouleOntologie\fm{a}$ and
$\recette_\ell\nDecouleOntologie\fm{a}$, 
\item the property $\fm{a}\Plus$ 
if $\recette_k\nDecouleOntologie\fm{a}$ and
$\recette_\ell\DecouleOntologie\fm{a}$, 
\item the property $\fm{a}\Eg$ 
if $\recette_k \DecouleOntologie\fm{a}$ and
$\recette_\ell\DecouleOntologie\fm{a}$. 
\end{itemize}
For example, if:
\begin{align*}
\recette_k &= \fm{chinese} \land \fm{soup} \land \ldots \land
\fm{peanut\_oil} \land \text{\emph{Nothing else}}\\
\recette_\ell &= \fm{chinese} \land \fm{soup} \land \ldots \land
\fm{olive\_oil} \land \text{\emph{Nothing else}} 
\end{align*}
then $\dRkl = \{ \fm{chinese}\Eg, \fm{soup}\Eg, \fm{oil}\Eg,
\fm{peanut\_oil}\Moins, \fm{olive\_oil}\Plus, \ldots\}$, provided that
$\fm{peanut\_oil} \DecouleOntologie \fm{oil}$,
$\fm{olive\_oil} \DecouleOntologie \fm{oil}$,
$\recette_\ell \nDecouleOntologie \fm{peanut\_oil}$ and
$\recette_k \nDecouleOntologie \fm{olive\_oil}$.

The inclusion relation $\estinclusdans$ constitutes a partial order on
$\LvarR$ that can be used to organize variations by generality: a variation
$\diffR$ is more general than a variation $\diffR'$ if $\diffR
\estinclusdans \diffR'$.

\paragraph{Mining.}
The learning process 
consists in highlighting some variations
$\diffR\in\LvarR$ that are more general than a  
``large'' number of elements 
 $\dRkl$ of $\trainingset$.
More formally, let
$$\supportFun{\diffR}
=
\frac
{
\card{\{\dRkl\in\trainingset\mid\diffR\estinclusdans\dRkl\}}
}
{\card{\trainingset}}
$$
Learning adaptation rules aims at finding the 
$\diffR\in\LvarR$ such that
$\supportFun{\diffR}\geq\seuilS$, where $\seuilS\in[0 ; 1]$ is 
a learning parameter called the support threshold.
It can be noticed that if $\diffR_1\estinclusdans\diffR_2$ then 
$\supportFun{\diffR_1}\geq\supportFun{\diffR_2}$.
%%VIR:debut
The support threshold also has an 
influence on the number of generated variations.
The number of generated variations increases when $\seuilS$ decreases.
Thus, specifying a high threshold restricts the generation of
variations to the most general ones, which can limit the number of 
generated variations and save computation time but has the effect to
discard the most specific ones from the result set.
%%VIR:fin

Each learned variation $\diffR = \{ \fm{p}_1, \fm{p}_2, \ldots
,\fm{p}_n \} \in \LvarR$ 
is interpreted as a substitution of the form
$\fm{A} \substituepar \fm{B}$ such that:
\begin{itemize}
\item $\fm{A} \DecouleOntologie \fm{a}$ 
  and $\fm{B} \nDecouleOntologie \fm{a}$
  if $\fm{a}\Moins\in\diffR$, 
\item $\fm{A} \nDecouleOntologie \fm{a}$ 
  and $\fm{B} \DecouleOntologie \fm{a}$
  if $\fm{a}\Plus\in\diffR$, 
\item $\fm{A} \DecouleOntologie \fm{a}$ 
  and $\fm{B} \DecouleOntologie \fm{a}$
  if $\fm{a}\Eg \in \diffR$.
\end{itemize}
For example, the variation 
$\diffR = \{ \fm{oil}\Eg, \fm{peanut\_oil}\Moins,
\fm{olive\_oil}\Plus \}
$ 
is interpreted as the substitution
$
\sigmaR = \fm{peanut\_oil} 
\substituepar \fm{olive\_oil}
$.
The conjunct $\fm{oil}$ is not present neither in $\fm{A}$ nor in
$\fm{B}$ since it is useless: $\fm{peanut\_oil} \DecouleOntologie
\fm{oil}$ and $\fm{olive\_oil} \DecouleOntologie \fm{oil}$.

\paragraph{Filtering.}
For a retrieved recipe $\sol(\source)$, the result set can be filtered in
order to retain only the substitutions 
$\sigmaR = \fm{A} \substituepar \fm{B}$ that can be applied to modify
$\sol(\source)$, i.e., such that $\sol(\source) \DecouleOntologie
\fm{A}$.
%%VIR:debut
%% For example, a substitution like $\fm{peanut\_oil} \land \fm{rice}
%% \substituepar \fm{olive\_oil}$ cannot be applied to modify the recipe
%% \emph{Wonton Soup} since the latter contains no rice as ingredient
%% (and thus $\sol(\source) \nDecouleOntologie \fm{peanut\_oil} \land
%% \fm{rice}$). 
%%VIR:fin

\paragraph{Validation.}
Knowledge discovery aims at building a model of reality from a set of
observations.
But as a model of a part of reality is only valid with respect to a  
particular observer, any learned substitution has to be validated
by a human expert in order to acquire the status of piece of knowledge.

%           --------------------------------------------
\subsection{Opportunistic Adaptation Knowledge Discovery}
%           --------------------------------------------

% quoi ?
The $\ca$ discovery process turns the case 
base into an additional source of adaptation knowledge.
This new source of knowledge is
used during a problem-solving session to provide the $\rapc$ system with 
adaptation knowledge ``on demand''.
% comment ?
A set of variations $\diffR$ is learned from the case base by
comparing two sets of recipes and 
each learned variation $\diffR$ is interpreted as a substitution
$\sigmaR$ that can be used to repair the adaptation path
$\adaptationPath$.
Each learned substitution $\sigmaR$ is presented
to the user for validation together with 
the corrected solution $\solCand{\cible}$ resulting from its
application. 
When the user validates the corrected solution, a new reformulation
$(\sigmaQ,\sigmaR)$ is added to the adaptation knowledge base $\akb$
so that the learned substitution $\sigmaR$ can be later reused to
adapt new recipes. 
% quand ?
The $\ca$ discovery process is triggered either during the adaptation
phase, to come up with suggestions of gradual
solution refinements (see section~\ref{sec:exemple1} for an example), 
or during the solution test phase to repair a failed adaptation in
response to the user's feedback (see section~\ref{sec:exemple2} for an
example). 

%           --------------
\subsection{Implementation}
%           --------------
To test the proposed adaptation knowledge acquisition method,
a prototype was implemented that integrates the
$\taaable$ system~\cite{taaable} and the $\cabamaka$
system~\cite{ijcai07}.
The case base contains $862$ recipes taken from the CCC 2008 recipe
set. 
The $\taaable$ system is used to perform retrieval and adaptation.
The $\cabamaka$ system is used to learn a set of substitutions $\sigmaR$ from
the case base from the comparison of two sets of recipes.
As in~\cite{ijcai07}, the mining step is performed thanks to a 
frequent closed itemset extraction algorithm.

%           -----------------------------------------
\subsection{A First Example: Cooking a Chocolate Cake}\label{sec:exemple1}
%           ----------------------------------------- 

An example is presented to illustrate how the case base is used as an
additional source of adaptation knowledge.
The $\ca$ discovery process is parametrized automatically and is used
to provide assistance to the user by 
suggesting some gradual refinements for the proposed solution.

\begin{enumerate}
\item \emph{Representing the Target Problem.}
In this example, the user wants to cook a chocolate
cake with baking chocolate and oranges.
The target problem is:
$$
\cible = \fm{cake} \land \fm{baking\_chocolate} \land \fm{orange}
$$
In the $\taaable$ interface, the field ``Ingredients I Want'' is
filled in with the tokens $\fm{baking\_chocolate}$ and $\fm{orange}$
and the field ``Types I Want'' is filled in with 
the token $\fm{cake}$.

\item \emph{Retrieval.}
The retrieval procedure generates the similarity path $\simPath =
\sigmaQ_1$ in which the substitution $\sigmaQ_1 =
\fm{baking\_chocolate} \substituepar \fm{chocolate}$ 
is generated automatically from the ontology $\Ontologie$ from the
axiom $\fm{baking\_chocolate} \limplique \fm{chocolate}$.
$\simPath$ is applied to $\cible$ in order to produce the modified
query $\source = \fm{cake} \land \fm{chocolate} \land \fm{orange}$.
The system retrieves the recipe \emph{Ultralight Chocolate Cake},
whose representation $\sol(\source)$ is:
$$
\sol(\source) = \fm{cake} \land \fm{cocoa} \land \fm{orange} \land
\ldots \land \text{\emph{Nothing else}}
$$
Since the ontology $\Ontologie$ contains the axiom $\fm{cocoa}
\limplique \fm{chocolate}$,  
$\sol(\source)$ solves the query $\source$:
$\sol(\source)$ is such that $\sol(\source) \DecouleOntologie
\source$.

\item \emph{Adaptation.}
$\akb$ is assumed to be empty, so to construct the adaptation path
$\adaptationPath$, the substitution $\fm{chocolate}
\substituepar \fm{baking\_chocolate}$ is generated automatically from
$\sigmaQ_1$. 
This substitution is further specialized into the
substitution $\sigmaR_1 = \fm{cocoa} \substituepar
\fm{baking\_chocolate}$, according to the axiom $\fm{cocoa} \limplique
\fm{chocolate}$ of $\Ontologie$.
A first solution $\solCand{\cible}$ is computed by applying to
$\sol(\source)$ the adaptation path $\adaptationPath = \sigmaR_1$.
The user 
%%is not satisfied with the proposed solution
suggests that an ingredient is missing in
$\solCand{\cible}$ but could not identify a repair strategy.
An $\ca$ discovery is triggered in order to suggest gradual
refinements of $\solCand{\cible}$.

\item \emph{Choosing the Training Set.}
The training set $\trainingset$ is chosen %%in an opportunistic manner
from $\sigmaR_1$: $\ca$ is learned by
comparing the recipes containing cocoa with the recipes containing
baking chocolate.
$\trainingset$ is composed of the set of
variations $\dRkl\in\LvarR$ between pairs of recipes 
$(\recette_k,\recette_\ell)\in\basedecas\times\basedecas$ such that
$
\{ \fm{cocoa}\Moins, \fm{baking\_chocolate}\Plus \} \estinclusdans \dRkl
$.

\item \emph{Mining and Filtering.}
A value is given to the support threshold $\seuilS$ and the mining
step outputs a set of variations.
A filter retains only the variations that correspond to substitutions
applicable to modify $\sol(\source)$.
%% This set of variations is filtered to retain only the variations that
%% correspond to substitutions applicable to modify $\sol(\source)$.

\item \emph{Solution Test and Validation.}
The user selects the learned variation\linebreak
$
\diffR = \{ \fm{cocoa}\Moins, \fm{baking\_chocolate}\Plus,
\fm{oil}\Moins\}
$ from the result set.
$\diffR$ is interpreted as the substitution
$
\sigmaR = \fm{cocoa} \land \fm{oil}
\substituepar \fm{baking\_chocolate}
$, which suggests to replace cocoa by baking
chocolate in the retrieved recipe and to remove oil.
The user explains this
rule by the fact that baking chocolate contains more fat than cocoa,
and therefore substituting cocoa by baking chocolate implies
to reduce the quantity of fat in the recipe.

Further solution refinements are proposed to the user.
The set of learned variations is filtered in order to retain only the
substitutions $\diffR'$ that 
are more specific than $\diffR$, i.e., such that $\diffR
\estinclusdans \diffR'$.
Among the retained variations is the variation 
$
\diffR' = \{ \fm{cocoa}\Moins, \fm{baking\_chocolate}\Plus,
\fm{oil}\Moins, \fm{vanilla}\Moins\}
$, which is interpreted as the substitution
$
\sigmaR' = \fm{cocoa} \land \fm{oil} \land \fm{vanilla}
\substituepar \fm{baking\_chocolate}
$.
$\sigmaR'$ suggests to also remove vanilla in the
recipe \emph{Ultralight Chocolate Cake}.
%
%\paragraph{Retaining the Learned Adaptation Knowledge.}
The user is satisfied with the refined solution $\solCand{\cible}$
resulting from the application of the adaptation path $\adaptationPath
= \sigmaR'$ to $\sol(\source)$, so the reformulation \linebreak
%%$(\sigmaQ,\sigmaR')$ 
{\small
$
(\fm{baking\_chocolate}\substituepar\fm{chocolate},
~\fm{cocoa} \land \fm{oil} \land \fm{vanilla} \substituepar
\fm{baking\_chocolate})
$}
is added to the adaptation knowledge base $\akb$.
%% , with:
%% \begin{align*}
%% \sigmaQ &= \fm{baking\_chocolate} \substituepar \fm{chocolate}\\
%% \sigmaR' &= \fm{cocoa} \land \fm{oil} \land \fm{vanilla} \substituepar
%% \fm{baking\_chocolate} 
%% \end{align*}
\end{enumerate}

%           ----------------------------------------
\subsection{A Second Example: Cooking a Chinese Soup}\label{sec:exemple2} 
%           ----------------------------------------

A second example is presented in which the $\ca$ discovery
process is triggered  in response to the user feedback in order to
repair the adaptation presented in Sect.~\ref{sec:taaable}. 
In this example, the user is encouraged to formulate the cause of the
adaptation failure.
A repair strategy is chosen that is used to parametrize the $\ca$ discovery 
process.

\begin{enumerate}
\item \emph{Representing the Target Problem.}
In this example, the target problem $\cible$ is:
$$
\cible = \fm{chinese} \land \fm{soup} \land \fm{leek} \land
\not\fm{peanut\_oil} 
$$
In the $\taaable$ interface, the field ``Ingredients I Want'' is
filled in with the token $\fm{leek}$, the field ``Ingredients I Don't
Want'' is filled in with the token $\fm{peanut\_oil}$ and the field
``Types I Want'' is filled in with the tokens $\fm{chinese}$ and
$\fm{soup}$. 

\item \emph{Retrieval.}
%The retrieval procedure is unchanged.
As in Sect.~\ref{sec:taaable}, two substitutions $\sigmaQ_1 =
\not\fm{peanut\_oil} \substituepar \emptyset$ and $\sigmaQ_2 =
\fm{leek} \substituepar \fm{onions}$ are generated automatically from
the ontology $\Ontologie$.
The similarity path $\simPath = \sigmaQ_2 \comp \sigmaQ_1$ is applied
to $\cible$ in order to produce 
the modified query $\source = \fm{chinese} \land \fm{soup} \land
\fm{onions}$. 
The system retrieves the recipe \emph{Wonton Soup}, whose
representation $\sol(\source)$ solves the query $\source$:
$\sol(\source)$ is such that
$\sol(\source) \DecouleOntologie \source$.  

\item \emph{Adaptation.}
Initially, $\akb = \emptyset$, so to construct the adaptation path
$\adaptationPath$, two substitutions $\sigmaR_1 = \emptyset
\substituepar \not\fm{peanut\_oil}$ and
$\sigmaR_2 = \fm{green\_onion} \substituepar \fm{leek}$ are
automatically generated from $\sigmaQ_1$ and $\sigmaQ_2$.

\item \emph{Solution Test and Validation.}
The solution $\solCand{\cible}$ is presented to the user for
validation, together with the adaptation path $\adaptationPath =
\sigmaR_1 \comp \sigmaR_2$ that was used to generate it.

\item \emph{The User is Unsatisfied!}
The user complains that the adapted recipe is practically unfeasible
because the proposed solution $\solCand{\cible}$
does not contain oil anymore, and oil is needed to saute the bok choy. 
% in response to the user's feedback.

\item \emph{What has Caused the Adaptation Failure?}
The cause of the adaptation failure is identified 
through interactions with the user.
The user validates the intermediate solution
$\solCand{\prb}$ that results from the application of the substitution
$\sigmaR_2 = \fm{green\_onion} \substituepar \fm{leek}$ to $\sol(\source)$.
But the user invalidates the solution $\solCand{\cible}$
that results from the application of $\sigmaR_1 = \emptyset
\substituepar \not\fm{peanut\_oil}$ to $\solCand{\prb}$.
The substitution $\sigmaR_1$ is identified as responsible for the
adaptation failure 
since its application results in the removal of oil in the
recipe.

\item \emph{Choosing a Repair Strategy.}
A repair strategy is chosen according to the user's feedback.
The user expresses the need for oil in the adapted
recipe, so the repair strategy consists in replacing peanut
oil by another oil.  
An $\ca$ discovery process is triggered to decide which oil to replace
peanut oil with.

\item \emph{Choosing the Training Set.}
%% The training set is chosen  opportunistically to reflect the
%% need for adaptation knowledge.
%% In order to decide which oil to replace peanut oil with, 
A set of recipes that contain peanut oil is compared 
with a set of recipes containing other types of oil. 
The training set $\trainingset$ is composed of the set of
variations $\dRkl\in\LvarR$ between pairs of recipes 
$(\recette_k,\recette_\ell)\in\basedecas\times\basedecas$ such that
$
\{ \fm{oil}\Eg, \fm{peanut\_oil}\Moins \} \estinclusdans \dRkl
$.
%%%   \begin{align*}
%%%     &\recette_k \DecouleOntologie \fm{oil} \land \fm{peanut\_oil}\\
%%%     &\recette_\ell \DecouleOntologie \fm{oil} \land \not\fm{peanut\_oil}
%%%   \end{align*} 

%% An optional filtering step may be included to further 
%% restrict the training set $\trainingset$ in order
%% to better take into account the problem-solving context.
%% For example, a filter may be applied to restrict the training set 
%% $\trainingset$ to asian recipes: the resulting training set
%% $\trainingset$ would be the set of  
%% variations $\dRkl\in\LvarR$ between pairs of recipes  
%% $(\recette_k,\recette_\ell)\in\basedecas\times\basedecas$ such that:
%% $$
%% \{ \fm{asian}\Eg, \fm{oil}\Eg, \fm{peanut\_oil}\Moins \} \estinclusdans \dRkl
%% $$
%% In the following, it is assumed that no filtering step was performed.

\item \emph{Mining and Filtering.}
A value is given to the support threshold $\seuilS$ and the mining
step outputs a set of variations.
A filter retains only the variations that correspond to substitutions
applicable to modify $\sol(\prb)$.

\item \emph{Solution Test and Validation.}
The user selects the learned variation\linebreak
$
\diffR = \{ \fm{oil}\Eg, \fm{peanut\_oil}\Moins,
\fm{olive\_oil}\Plus\}
$ from the result set.
$\diffR$ is interpreted as the substitution
$
\sigmaR = \fm{peanut\_oil}
\substituepar \fm{olive\_oil}
$, which suggests to replace peanut oil by olive oil in the
retrieved recipe. 
The adaptation path $\adaptationPath = \sigmaR \comp \sigmaR_2$ is
computed and the repaired solution $\solCand{\cible}$
%% $$ \solCand{\cible} = 
%% \fm{chinese} \land \fm{soup} \land \fm{leek} \land \ldots \land
%% \fm{olive\_oil} \land \text{\emph{Nothing else}} 
%% $$
is presented to the user for validation.
%
%\paragraph{Retaining the Learned Adaptation Knowledge.}
The user is satisfied with the corrected solution $\solCand{\cible}$, so
the reformulation 
%%$(\sigmaQ,\sigmaR)$ 
$
(\emptyset \substituepar \not\fm{peanut\_oil},
~\fm{peanut\_oil} \substituepar \fm{olive\_oil})
$
is added to the adaptation knowledge base $\akb$.
%% , with:
%% \begin{align*}
%% \sigmaQ &= \emptyset \substituepar \not\fm{peanut\_oil}\\
%% \sigmaR &= \fm{peanut\_oil} \substituepar \fm{olive\_oil}
%% \end{align*}
\end{enumerate}

%<       ---------------------------
\section{Discussion and Related Work}\label{sec:relatedwork}
%<       ---------------------------

% Related Work: Fadi
$\ca$ acquisition is a difficult task
that is recognized to be a major bottleneck for $\rapc$ system designers
due to the high knowledge-engineering costs it generates.  
To overcome these knowledge-engineering costs, a few approaches (e.g., 
\cite{ijcai07,craw06,hanney97})
have applied machine learning techniques to learn $\ca$
offline from differences between cases of the case base.
In \cite{hanney97}, a set of pairs of source cases is
selected from the case base and each selected pair of source cases is
considered as a specific adaptation rule.
The featural differences between problems 
constitute the antecedent part of the rule and the featural
differences between 
solutions constitute the consequent part.
Michalski's closing interval rule algorithm 
%~\note{ref:Michalski83} 
is then applied to generalize adaptation rule antecedents. 
In \cite{craw06},  adaptation knowledge takes the form of a set of
adaptation cases. 
Each adaptation case associates an adaptation action to a
representation of the 
differences between the two source problems.
Machine learning algorithms like C4.5 or RISE are
applied to learn generalized adaptation knowledge from these
adaptation cases
in order to improve the system's case-based adaptation procedure.
%% In this approach, adaptation is case-based and aims at refining a
%% retrieved solution. 
%
%% In \cite{DBLP:journals/ci/ShiuYSW01}, 
%% a set of general adaptation rules is learned from pairs of similar
%% source cases using a fuzzy decision tree algorithm.
%% The approach was successfully applied for case base maintenance:
%% the authors showed that the learned rules can be used to generate the
%% whole case base from a set of representative cases.
%% Case knowledge is thus transferred to adaptation knowledge in order to
%% reduce the size of the case base.

% discussion: TS
When applying machine learning techniques to learn adaptation
knowledge from differences between cases, one main challenge concerns
the choice of the training set: which cases are worth comparing?
Arguing that 
(1)~the size of the training set should be reduced to
minimize the cost of the adaptation rule generation process and that
(2)~the source cases that are worth comparing should be the ones 
that are more similar,
only the pairs of source cases that were judged to be similar
according to a given similarity measure are selected in \cite{craw06}
and \cite{hanney97}. 
However, committing to a
particular similarity measure might be somewhat arbitrary.
Therefore, in \cite{ijcai07}, the authors decided to include in the
training set all  
the pairs of distinct source cases of the case base.
This paper introduces a third approach: the choice of the training set is
determined interactively and according to the problem-solving context,
taking advantage of the fact that the $\ca$ discovery process is
triggered on-line.  
This approach appears to be very promising since the learning
algorithm can be parametrized in order to learn only the knowledge that
is needed to solve the target problem.

% discussion: complex strategies
The examples presented above also show that knowledge discovery
techniques allow to come 
up with more complex adaptation strategies 
than the simple one-to-one ingredient substitutions generated from the
ontology $\Ontologie$.
In particular, these techniques can help identifying interactions
between the different ingredients that appear in the recipes (like e.g.,
that cocoa contains less fat than baking chocolate, so oil should be
removed) as well as co-occurrences of ingredients
(like say, that cinnamon is well-suited with apples).
Besides, adaptation knowledge is learned at different levels of
generality, so the user can be guided into gradual solution
refinements.  

% partie Amelie
Several $\rapc$ systems make use of interactive and/or opportunistic knowledge
acquisition approaches to improve their learning capabilities.
For example, in Creek, an approach that combines case-based and model-based
methods, general knowledge is acquired through interactions with the
user~\cite{aamodt2004a}.
This knowledge acquisition process is provided in addition to the
traditional case acquisition and allows the system to acquire knowledge
that cannot be captured through cases only.
In the Dial system, adaptation knowledge is acquired in the form of
adaptation cases: when a case has to be adapted, the adaptation process is
memorized in the form of a case and can be reused to adapt another case.
Hence, adaptation knowledge is acquired through a $\rapc$ process inside the
main $\rapc$ cycle.
It must be remarked that adaptation cases can either be built automatically
by adaptation of previous adaptation cases or manually by a user who
interactively builds the adaptation case in response to a problem by
selecting the appropriates operations to perform \cite{leake1996b}.
Hence, knowledge acquisition in Dial appears to be both interactive and
opportunistic.
Chef is obviously related to the work described here \cite{hammond1986a}.
Chef is a case-based planner in the cooking domain, its task is to build
recipes on the basis of a user's request.
The input of the system is a set of goals (tastes, textures, ingredients,
types of dishes) and the output is a plan for a single recipe that satisfies
all the goals.
To solve this task, Chef is able to build new plans from old ones stored in
memory.
The system is provided with the ability to choose plans on the basis of the
problems that they solve as well as the goals they satisfy, but it is also
able to predict problems and to modify plans to avoid failures (plans are
indexed in memory by the problems they avoid).
Hence, Chef learns by providing causal explanations of failures thus marking
elements as "predictive" of failures.
In other words, the acquired knowledge allows the system to avoid identical
failures to occur again.
In our approach, we propose to go one step further by using failure to
acquire knowledge that can be more widely used.

%<       --------------------------
\section{Conclusion and Future Work}\label{sec:conclusion}
%<       --------------------------

In this paper, a novel approach for adaptation knowledge acquisition 
is presented in which the knowledge learned at problem-solving time
by knowledge discovery techniques is directly reused for problem-solving.
%% The proposed approach combines two previous methods:
%% one that learns $\ca$ from differences between cases
%% and another that acquires $\ca$ at problem-solving time through
%% interactions with the user.
An application is proposed in the context of the cooking $\rapc$
system $\taaable$ and the feasibility of the approach is demonstrated
on some use cases.
Future work will include developing a graphical user interface and doing
more extensive testing.  
% role de l'utilisateur/expert.
Opportunistic and interactive knowledge discovery in $\taaable$
implies that the user plays the role of the domain expert, which raises
several issues. 
For example, how to be sure that the knowledge expressed by a
particular user is valuable? How to ensure that the adaptation
knowledge base will remain consistent with time?
Besides, $\taaable$ is meant to be multi-user, so if the system's
knowledge evolves with experience, some synchronization problems might
occur. 
Therefore, the envisioned multi-user, ever-learning $\taaable$ system
needs to be 
thought of as a collaborative tool in which knowledge acquired by some
users can be revised by others.

%\note{travailler sur les quantites}

%<---------------->%
% LA BIBLIOGRAPHIE %
%<---------------->%

\end{document}